\documentclass[10pt,twocolumn,letterpaper]{article}

\usepackage{iccv}
\usepackage{times}
\usepackage{epsfig}
\usepackage{graphicx}
\usepackage{amsmath}
\usepackage{amssymb}
\usepackage[table,xcdraw]{xcolor}
\usepackage{float}
\usepackage{enumitem}
\usepackage[accsupp]{axessibility}  
\setitemize[1]{itemsep=0pt,partopsep=0pt,parsep=\parskip,topsep=0pt}

\usepackage{multirow,xcolor}
\usepackage{booktabs}
\usepackage{threeparttable}
\usepackage{bbm}
\usepackage{comment}
\newcommand{\MEAAD}{\texttt{\textbf{MEAAD}}\xspace}

\usepackage[pagebackref=true,breaklinks=true,letterpaper=true,colorlinks,bookmarks=false]{hyperref}

\iccvfinalcopy 

\def\iccvPaperID{9788} 

\ificcvfinal\fi

\usepackage[T1]{fontenc}
\usepackage[utf8]{inputenc}
\usepackage{authblk}

\title{Multi-Expert Adversarial Attack Detection in Person Re-identification Using Context Inconsistency}
\author[1,2]{Xueping Wang}
\author[3]{Shasha Li}
\author[1,2]{Min Liu \thanks{Corresponding author: liu\_min@hnu.edu.cn. 

X. Wang was a visiting student at UCR in 2019-20.}}
\author[1,2]{Yaonan Wang}
\author[3]{Amit K. Roy-Chowdhury}
\affil[1]{College of Electrical and Information Engineering, Hunan University, China}

\affil[2]{National Engineering Laboratory for Robot Visual Perception and Control Technology, China}

\affil[3]{University of California, Riverside}

\begin{document}

\maketitle
\thispagestyle{empty}

\begin{abstract}
\textcolor{black}{The success of deep neural networks (DNNs) has promoted the widespread applications of person re-identification (ReID). 
However, ReID systems inherit the vulnerability of DNNs to malicious attacks of visually inconspicuous adversarial perturbations. 
Detection of adversarial attacks is, therefore, a fundamental requirement for robust ReID systems.
In this work, we propose a Multi-Expert Adversarial Attack Detection (\MEAAD) approach to achieve this goal by checking context inconsistency, which is suitable for any DNN-based ReID systems.
Specifically, three kinds of context inconsistencies caused by adversarial attacks 
are employed to learn a detector for distinguishing the perturbed examples, i.e., a) the embedding distances between a perturbed query person image and its top-K retrievals are generally larger than those between a benign query image and its top-K retrievals, b) the embedding distances among the top-K retrievals of a perturbed query image are larger than those of a benign query image, c) the top-K retrievals of a benign query image obtained with multiple expert ReID models tend to be consistent, which is not preserved when attacks are present.
Extensive experiments on the Market1501 and DukeMTMC-ReID datasets show that, as the first adversarial attack detection approach for ReID, \MEAAD effectively detects various adversarial attacks and achieves high ROC-AUC (over 97.5\%).} 


\end{abstract}

\section{Introduction}
The success of DNNs has benefited a wide range of computer vision tasks, such as image classification \cite{he2016resnet,krizhevsky2012imagenet}, 
object detection \cite{he2017mask,ren2015faster},
face recognition \cite{rao2017learning,li2020measurement},
video classification \cite{li2018adversarial,zhang2020motion}, and person ReID \cite{zheng2019pyramidal,li2018hacnn,hermans2017defense,zheng2019pyramidal,qian2017mudeep}. Person ReID is a critical task aiming to retrieve pedestrians across multiple non-overlapping cameras. By learning the discriminative feature embedding and adaptive distance metric models, DNNs-based ReID models, in recent years, have extensive applications in video surveillance or criminal identification for public safety.
However, recent research has found that these models inherit the vulnerability of DNNs to adversarial examples  \cite{wang2020deepmisrank,wang2019advpattern,bai2020adversarial,ding2019universal} which are slightly perturbed input images but lead DNNs to make wrong predictions \cite{goodfellow2015fgsm,szegedy2013intriguing}. 
Detection of adversarial examples is, therefore, a fundamental requirement for robust ReID systems because the insecurity of ReID systems may cause severe losses.
However, ReID is defined as a ranking problem rather than a classification problem and thus existing defense methods for image classification \cite{cohen2020detecting,kurakin2016adversarial,papernot2018deep,ma2018characterizing,miyato2015distributional,tramer2017ensemble} do not fit the person ReID problem.

\begin{figure*}
\centerline{\includegraphics[width=1.7\columnwidth]{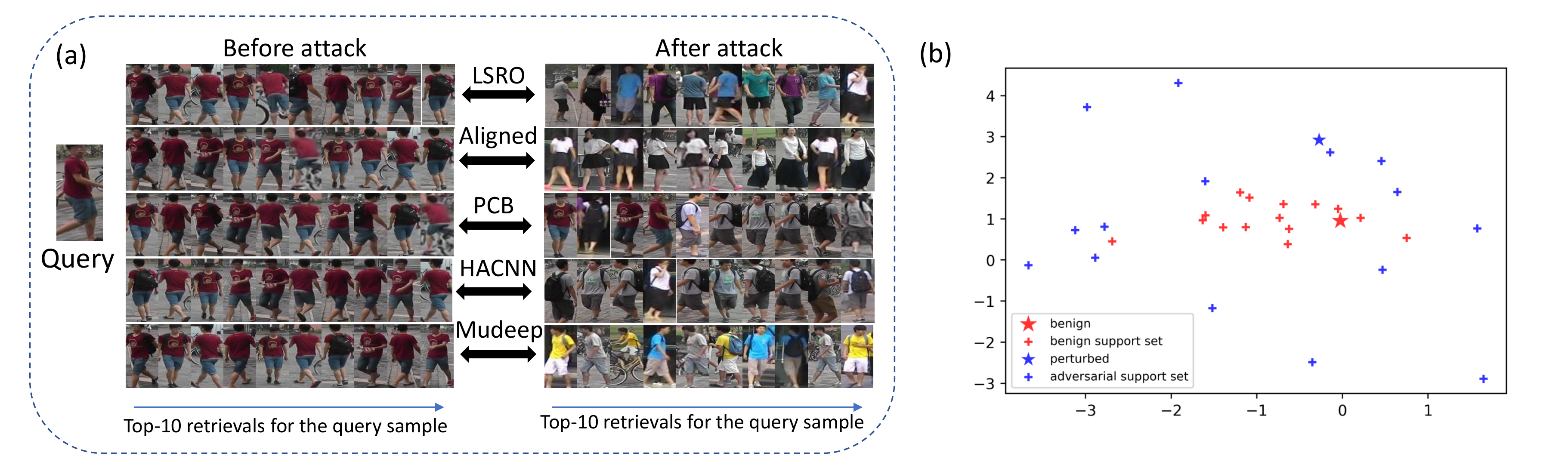}}
\caption{(a) shows the top-10 retrievals for a query sample before and after an adversarial attack. Five state-of-the-art person ReID models, i.e., LSRO \cite{zheng2017lsro}, AlignedReID \cite{zhang2017alignedreid}, PCB \cite{sun2018pcb}, HACNN \cite{li2018hacnn} and Mudeep \cite{qian2017mudeep} are used as the expert models. Deep Mis-Ranking attack \cite{wang2020deepmisrank} is used to generate the adversarial perturbations and AlignedReID is the attack target model. The top-10 retrievals of a benign query (before attack) are consistent across multiple expert models, while they are messy for a perturbed query sample (after attack). (b) presents (using PCA) the embedding space of an expert model (AlignedReID). The original query sample is marked with red star and its retrievals are marked with red plus marker. The perturbed query sample is marked with blue star and its retrievals are marked with blue plus marker. We observe that the retrievals of the benign query sample gather tightly around the benign query sample in the embedding space. In comparison, the retrievals of the perturbed query sample spread over the space. We have quantitative and more detailed results in Fig.~\ref{fig:fig2}.}

\label{fig:fig1}
\vspace{-0.4cm}
\end{figure*}

In addition, the top-$K$ retrievals output by person ReID systems, compared to the prediction label in classification task, contain richer information and can be potentially employed to detect adversarial attacks. To illustrate, let's consider the top-10 retrievals obtained with five different state-of-the-art person ReID systems (LSRO \cite{zheng2017lsro}, AlignedReID \cite{zhang2017alignedreid}, PCB \cite{sun2018pcb}, HACNN \cite{li2018hacnn} and Mudeep \cite{qian2017mudeep}) of a query sample before and after an adversarial attack in Fig.~\ref{fig:fig1}(a). \emph{When considering the retrievals returned by a single ReID system, e.g. LSRO, they are visually more similar to the query image before attack than that after attack, and they are visually more similar to each other before attack. When considering the retrievals returned by different expert models, they are consistent before attack but certainly not after attack.} 
We did an empirical study as shown in Fig.~\ref{fig:fig1}(b) and found that the embedding distance is able to reflect visual similarity; specifically, the retrievals of the benign query tend to gather together in the embedding space while the retrievals of the perturbed query tend to spread over.

Inspired by these observations, we propose Multi-Expert Adversarial Attack Detection which detects adversarial attacks for person ReID systems by detecting context inconsistency. 
To the best of our knowledge, this is the first strategy to detect adversarial attacks against person ReID systems.
To make use of the heterogeneity of different ReID models, 
we use multiple ReID networks with different architectures as expert models in \MEAAD.  We define support set as the top-$K$ retrievals output by a single expert model.
Context used in \MEAAD accounts for three types of relations: 1) the relations between the query and its support samples returned by a single expert ({\it Query-Support Affinity}); 2) the relations among the support samples returned by a single expert ({\it Support-Support Affinity}); 3) the relations between the support samples returned by one expert and those returned by another ({\it Cross-Expert Affinity}).
We then train a detector with the context features of both benign and perturbed query samples and use it to detect adversarial attacks during testing. The contributions are as below,
\begin{itemize}[leftmargin=*]
    \item To the best of our knowledge, this is the first adversarial attack detection strategy for the defense of ReID systems. 
    \item We empirically study the side effect brought by adversarial attacks, i.e., context inconsistency in the retrieval results. We then propose \MEAAD which aims to detect adversarial attacks by checking context inconsistency of a query sample to be detected. 
    \item Extensive experiments on the Market1501 and DukeMTMC-ReID datasets show that, \MEAAD effectively detects various adversarial attacks and achieves high ROC-AUC (over 97.5\% in all cases).
\end{itemize}

\begin{figure*}
\centerline{\includegraphics[width=1.85\columnwidth]{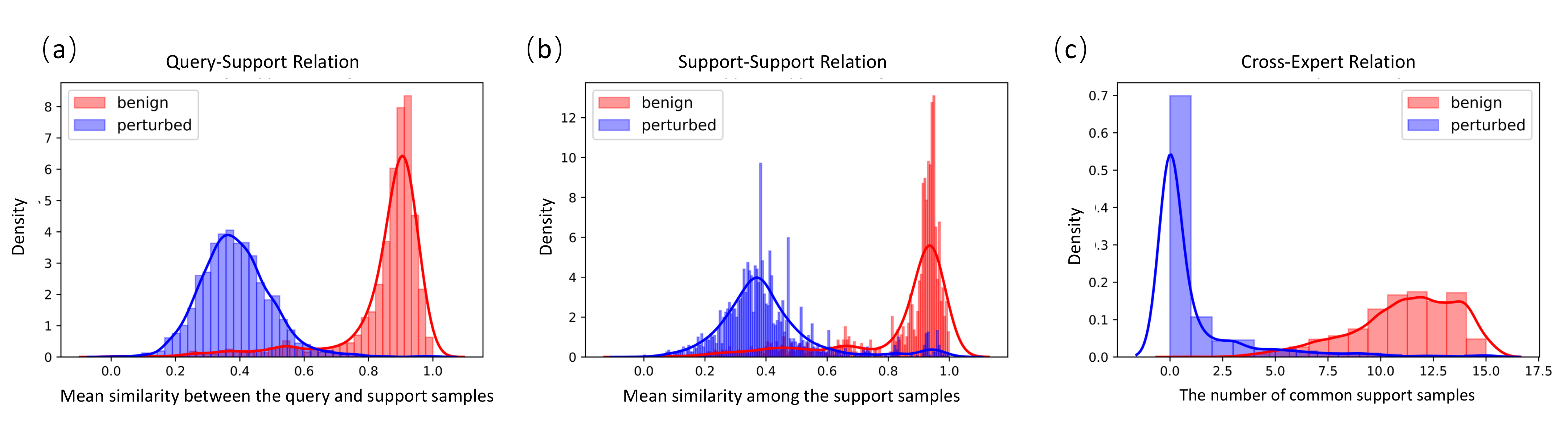}}
\caption{Empirical study results: (a) plots the query-support relation distribution for both benign and perturbed query samples. The query-support relation is defined to be the average of cosine similarity between the embedding feature of the query sample and the embedding features of the support samples. (b) plots the support-support relation distribution. The support-support relation is defined to be the average of cosine similarity among the embedding features of the support samples for each query image in the same support set.
(c) plots the cross-expert relation distribution. We use the same number of support samples across all support sets to describe cross-expert relation. 
} 
\label{fig:fig2}
\vspace{-0.4cm}
\end{figure*}

\section{Related works}

\subsection{Person re-identification}
\label{sec:reID}
Person re-identification is a cross-camera instance retrieval problem, which aims at searching persons across multiple cameras.
With the advancement of deep learning, person ReID has achieved inspiring performance on the widely used benchmarks. In this field, deep learning-based feature representation methods which focus on developing the feature construction strategies  have been widely used \cite{zhang2017alignedreid,li2018hacnn,qian2017mudeep}. 
In \cite{hermans2017defense,zheng2019pyramidal,xueping2020tcsvt}, the authors proposed to employ deep metric learning models to address the person ReID task, which aim at designing the training objectives with different loss functions or sampling strategies. In recent years, using GAN to transfer the source domain images to target-domain style is a popular approach for ReID \cite{deng2018spgan,zhong2018camerastyle,zhong2018hhl}. With the generated images, this enables using supervised ReID models in the unlabeled target domain. Another direction is to learn ReID models from limited labeled data \cite{raychaudhuri2020exploiting,xueping2021tip}.
These methods have achieved impressive performance. 
In our framework, we adopt the state-of-the-art person ReID models with different network architectures as our expert models and extract context features from the outputs of these experts for adversarial attack detection.

\subsection{Adversarial attacks}
Adversarial attacks have achieved remarkable success in fooling DNN-based systems, e.g., image classification \cite{goodfellow2015fgsm,li2020deeptrack,kurakin2016bim,madry2017pgd,moosavi2016deepfool,papernot2016cw,li2018adversarial}
and object detection \cite{chen2018shapeshifter,zhao2019seeing}, etc. A few adversarial attacks have been proposed for attacking ReID models. Wang et al. \cite{wang2020deepmisrank} proposed a learning-to-misrank formulation to perturb the ranking of the ReID system outputs.
Ding et al. \cite{ding2019universal} proposed an effective method to train universal adversarial perturbations (UAPs) against person ReID models from the global list-wise perspective. \cite{bouniot2020vulnerability, bai2020adversarial} proposed adversarial metric attack to perturb the ReID systems. Instead of the previous digital perturbations, Wang et al. \cite{wang2019advpattern} implemented robust physical-world attacks against deep ReID for generating adversarial patterns on clothes, which learns the variations of image pairs across cameras to pull closer the image features from the same camera, while pushing features from different cameras farther. Our defense strategy is dependent on the contextual information, and therefore does not rely heavily on the mechanism to generate the perturbations.

\subsection{Adversarial defense}
To address the vulnerability of DNNs to adversarial attacks, some adversarial training-based defense approaches have been proposed \cite{goodfellow2015fgsm,kurakin2016adversarial,miyato2015distributional,tramer2017ensemble}.
However, adversarial training-based defense methods degrade the natural performance of the target models and they can be evaded by the optimization-based attack \cite{carlini2017adversarial}, either wholly or partially. 
Recent works have focused on detection-based defense methods which aim at distinguishing adversarial examples from benign ones \cite{liu2019detection,li2020connecting,papernot2018deep,cohen2020detecting,ma2018characterizing}.
Li et al. \cite{li2020connecting} proposed a context inconsistency-based adversarial perturbation detection method for object detection task. They constructed a fully connected graph for each detected object by accounting for four types of region relationships, and trained a classifier for each category. Yin et al. recently presented how to use language descriptions to detect adversarial attacks through context inconsistencies \cite{yin2021adversarial}.
\cite{liu2019detection} applied steganalysis techniques to model the dependence between adjacent pixels; adversarial perturbations in most cases alter the dependence between pixels and thus can be detected by their method.
However, most of these methods are developed for the classification task and they are not suitable for defending the person ReID models (ranking systems), because in a classification task, the training and testing set share the same categories, while in ReID, there is no category overlap between them.

\section{Methodology}
\subsection{Threat model}

The attacker's goal is to cause the target ReID system to retrieve person images of wrong identities. 
In this paper, we assume the attacker is able to launch attacks against the target ReID system by perturbing the query images, but not poisoning the gallery images. 
The same threat model has been used in \cite{wang2020deepmisrank,liu2019s,ding2019universal,bouniot2020vulnerability,zheng2018open,tolias2019targeted}, and is reasonable as galleries are very large (usually secured) and attacking a large number of gallery images is very time-consuming \cite{zheng2018open}.
We assume the attack target ReID model is white-box to the attacker
because even through in practical attacks it could be black-box, works~\cite{papernot2017practical,dong2019efficient} have shown that  attackers could estimate the function of a black-box model by making queries and reasoning on the query results. This assumption favors the attacker, 
and thus makes the defense systems more robust.
The other expert models used for consistency check never output retrieval results to the users, thus we assume the attacker is not aware of their existence. Note that we also explore two adaptive attacks where the attacker is aware of all experts and our defense scheme in Section~\ref{adaptive attack}.

\subsection{Empirical study on context inconsistency}
While the adversarial examples fool the ReID system to retrieve wrong images from the gallery set, they have a side effect, i.e., causing "messy" retrieval results as shown in Fig.~\ref{fig:fig1}. We define three relations to describe ``messy". Before introducing the three relations, we define the top-$K$ retrievals returned by the ReID system as the support set and each retrieval in it as the support sample. We refer to the support set of a benign query sample as benign support set and refer to that of a perturbed query sample as adversarial support set for simplicity. The empirical study is done with 2,000 benign query samples and 2,000 perturbed query samples obtained with the Deep Mis-Ranking attack \cite{wang2020deepmisrank} and top-15 retrievals of each query are used as the support set. Two person ReID systems (LSRO \cite{zheng2017lsro} and AlignedReID \cite{zhang2017alignedreid}) are used in the empirical study, and we call each system an expert model.

\noindent \textbf{Query-Support Relation.}
The retrieved images in the benign support set tend to be similar to the query image. We validate whether the embedding feature similarity between the query image and the support set reflects the same trend and the results are shown in Fig.~\ref{fig:fig2}(a). We define Query-Support Relation as the average of cosine similarity between the embedding feature of the query sample and the embedding features of the support samples. We observe that compared to benign query samples, perturbed queries generally have lower similarity to its support samples in the embedding space. This implies that we could distinguish benign and attack by the Query-Support Relation.

\noindent \textbf{Support-Support Relation.}
The retrieved images in the benign support set tend to be similar to each other. We validate whether the embedding feature similarity among the support samples reflects the same trend and the results are shown in Fig.~\ref{fig:fig2}(b). We define Support-Support Relation as the average of cosine similarity among the embedding features of the support samples for each query image. We observe that compared to benign support samples, adversarial support samples have lower similarity to each other in the embedding space. This implies that we could distinguish benign and attack by the Support-Support Relation.

\noindent \textbf{Cross-Expert Relation.}
We observe from Fig.~\ref{fig:fig1} that the benign support sets returned by different expert models overlap with each other a lot. 
We use the number of the common support samples returned by all expert models to describe Cross-Expert Relation. Fig.~\ref{fig:fig2}(c) shows that for a benign query sample, different expert models tend to return the same retrievals, which implies that the Cross-Expert Relation could be used to distinguish benign and attack. 

\begin{figure*}
\centerline{\includegraphics[width=1.7\columnwidth]{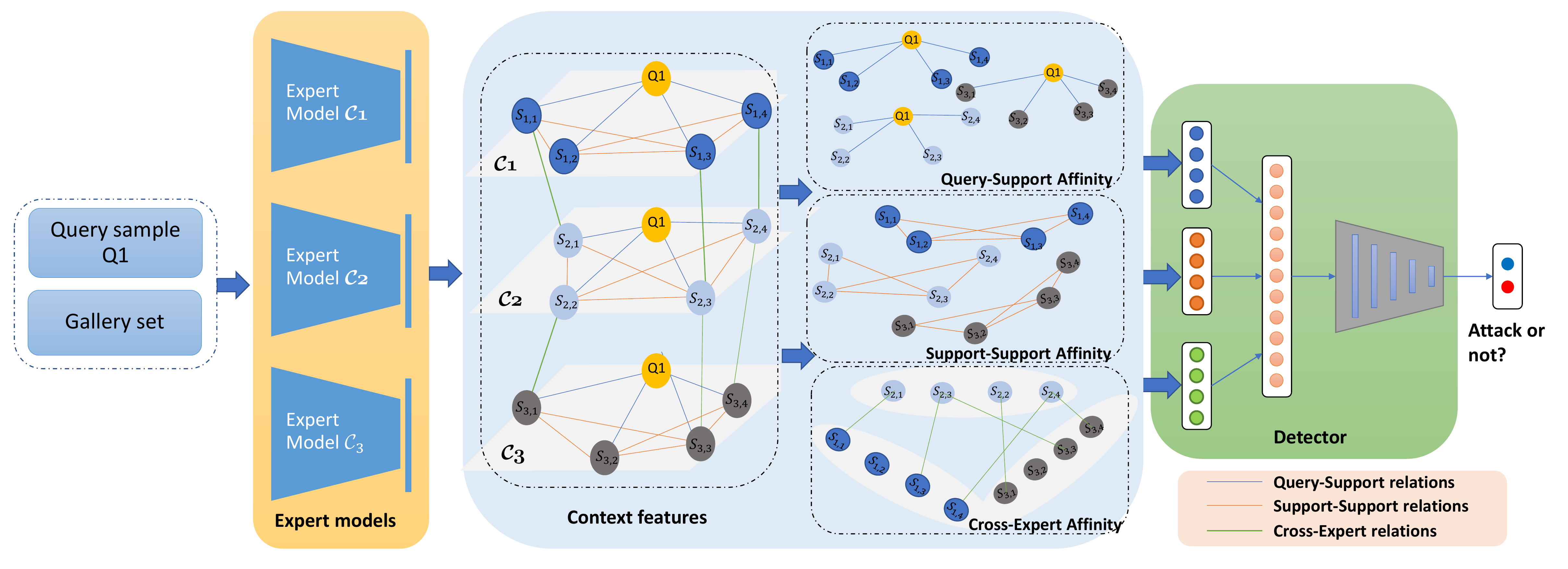}}
\caption{The pipeline of the proposed Multi-Expert Adversarial Attack Detection system. We employ multiple state-of-the-art ReID networks with different architectures as the expert models. The top-$K$ retrievals of a query sample are defined as a support set and each retrieval is a support sample. Based on query-support affinity, support-support affinity and cross-expert affinity, we define context feature for each query image and its support sets. A detector with context features as input is then learnt to distinguish attack from benign. It may be noted that in this figure we use three expert models and top-4 retrievals as an example to illustrate our framework.}
\label{fig:fig3}
\vspace{-0.4cm}
\end{figure*}

\subsection{Multi-expert adversarial attack detection}
Inspired by the above empirical studies, we propose  multi-expert adversarial attack detection illustrated in Fig.~\ref{fig:fig3} to distinguish the perturbed samples from benign ones by checking context inconsistency of the query samples. 

\subsubsection{Context feature}
To formulate the problem, we use $I$ to denote the query image and use $F_i(\cdot),i=1,2,...,N$ to denote the functions of the $N$ expert models. We denote the support set (top-$K$ retrievals) retrieved by the $i^{th}$ expert model as $\mathbf{S}_i=\{S_{i,j} | j=1,..K\}$. Each model learns a mapping from the image space to its latent feature embedding space. Therefore the embedding feature of $I$ with the $i^{th}$ expert model can be represented as $F_i(I)$. The heterogeneity of these expert models provides multi-view information for each query sample. The context feature is composed of three parts: query-support affinity, support-support affinity and cross-expert affinity, and we describe each in details. 

\noindent \textbf{Query-Support Affinity.} 
We extract the query-support affinity feature for each expert model in the same way. Therefore, we use $\mathbf{S} =\{S_{j} | j=1,..K\} $ instead of $\mathbf{S}_i = \{S_{i,j} | j=1,..K\}$ for simplicity afterwards. Similarly, we use $F(\cdot)$ instead of $F_i(\cdot)$.
If we use $A_{q-s}$ to denote the query-support affinity feature for the current expert model, then $A_{q-s}$ is a vector of $K$ dimension and the $j^{th}$ element is defined as the cosine similarity between the embedding feature of $I$ and the embedding feature of the support sample $S_j$ as shown in Eq.~\ref{equ:q-s}.
\begin{equation}
\label{equ:q-s}
    A_{q-s}[j] = {CosSimilarity}( F(I), F(S_j) )
\end{equation}
We calculate $A_{q-s}$ for all the expert models and stack them together, which is the final query-support affinity feature with dimension $N*K$.

\noindent \textbf{Support-Support Affinity.}
We extract support-support affinity feature for each expert model in the same way. If we use $A_{s-s}$ to denote the support-support affinity feature for the current expert model, then $A_{s-s}$ is a matrix of $K*K$ dimension and the element on $(i,j)$ is defined as the cosine similarity between the embedding feature of $S_i$ and the embedding feature of $S_j$ as shown in Eq.~\ref{equ:s-s}.
\begin{equation}
\label{equ:s-s}
    A_{s-s}[i,j] = {CosSimilarity}( F(S_i), F(S_j) )
\end{equation}
Note that $A_{s-s}$ is a symmetric matrix and the diagonal elements are always 1 (suppose the embedding feature is normalized). Therefore, instead of keeping all the elements, we keep the $K*(K-1)/2$ elements in the upper-right (or lower-left) matrix and $A_{s-s}$ becomes a vector of $K'=K*(K-1)/2$ dimension.
We calculate $A_{s-s}$ for all the expert models and stack them together, which is the final support-support affinity feature with dimension $N*K'$.

\noindent \textbf{Cross-Expert Affinity.}
To calculate the cross-expert affinity, we need the support sets of all the $N$ expert models. At each time, we choose an expert model as the base model and other $N-1$ expert models are called member models. We choose the base model in turns, and thus if we use $A_{c-e}$ to denote the cross-expert affinity feature then $A_{c-e}$ is a matrix and the $i^{th}$ row of the matrix is the feature calculated when the $i^{th}$ expert model is chosen as the base model. The element on $(i,j)$ is defined as the frequency that the $j^{th}$ support sample output by the base model (denoted as $S_{i,j}$) appears in the support sets output by the member expert models as shown in Eq.~\ref{equ:c-e}.
\begin{equation}
\label{equ:c-e}
    A_{c-e}[i,j] = \frac{\sum_{l \in \{1,..,N\}-\{i\}}\mathbbm{1}(S_{i,j} \in \mathbf{S}_{l})}{N-1}
\end{equation}
$\mathbbm 1(\cdot)$ is an indicator function which gives a value of 1 when the argument is true. 
Therefore, the cross-expert affinity feature is a matrix with dimension $N*K$.

In summary, there are three parts of the context feature, i.e.,  query-support affinity feature $A_{q-s} \in \mathbb R^{N*K}$, support-support affinity feature $A_{s-s}\in \mathbb R^{N*K'}$ and cross-expert affinity feature $A_{c-e}\in \mathbb R^{N*K}$.  We flatten all the matrices into vectors and concatenate them together as the final context feature for one query sample. We use $x$ to denote the context feature, thus $x \in \mathbb R^{d}, d=N*K+N*K'+N*K$.
\subsubsection{Adversarial attack detector} 
With the context feature defined, the next step is to learn an adversarial attack detector. As shown in Fig.~\ref{fig:fig3}, the detector is basically a binary classifier which takes context features as inputs and outputs whether the query image is perturbed or not.  
Since the input is of relative low dimension,
we use Multi-Layer Perceptron (MLP) classifier as the detector. 

To train the detector, we extract context features with benign query samples and assign classification label $y=0$ to them, and also extract context features with perturbed query samples and assign classification label $y=1$ to those context features. Therefore, the training set is $\{(x_i,y_i)| i=1,2,..M \}$ and $M$ is the size of the training set. During testing, given a query sample, we first extract the context feature from the query sample and its support sets retrieved by multiple expert models, and then input the context feature into the detector to decide if the query sample is perturbed. 

\begin{table*}[t]
\centering
\caption{Comparison with the state-of-the-art adversarial attack detection methods on the Market1501 and DukeMTMC-ReID datasets against Deep Mis-Ranking and advPattern attacks.}
\small
\setlength{\tabcolsep}{8pt}
\begin{threeparttable}
\begin{tabular}{lllllllllllll}
\toprule
\multicolumn{1}{c}{\multirow{3}{*}{Defense Methods}} & \multicolumn{6}{c}{Market1501}                                                                                                                                     & \multicolumn{6}{c}{DukeMTMC-ReID}                                     \\ \cline{2-13}
\multicolumn{1}{c}{}                         & \multicolumn{3}{c|}{Deep Mis-Ranking}                                             & \multicolumn{3}{c|}{advPattern}                                                & \multicolumn{3}{c|}{Deep Mis-Ranking} & \multicolumn{3}{c}{advPattern} \\ \cline{2-13}
\multicolumn{1}{c}{}                         & \multicolumn{1}{l}{Acc}  & \multicolumn{1}{l}{AUC}  & \multicolumn{1}{l|}{F1}  & \multicolumn{1}{l}{Acc} & \multicolumn{1}{l}{AUC} & \multicolumn{1}{l|}{F1} & Acc        & AUC        & \multicolumn{1}{l|}{F1}        & Acc      & AUC      & F1      \\ \cline{1-13}

\multicolumn{1}{l}{D$k$NN \cite{papernot2018deep}}                    & \multicolumn{1}{l}{87.9} & \multicolumn{1}{l}{93.8} & \multicolumn{1}{l|}{86.9} & \multicolumn{1}{l}{98.9}    & \multicolumn{1}{l}{99.1}    & \multicolumn{1}{l|}{99.0}    &90.2  &97.8 &\multicolumn{1}{l|}{91.1}    &98.5   &99.0   &98.6 \\ 

LID \cite{ma2018characterizing}& 90.6   &96.2  &\multicolumn{1}{l|}{91.1}  &99.3  &99.7  &\multicolumn{1}{l|}{99.4}  &87.4  &95.2   &\multicolumn{1}{l|}{88.1}  &99.4  &99.6  &\bf 99.7  \\ 
SRM \cite{liu2019detection}& 94.3   &98.2  &\multicolumn{1}{l|}{94.1}  &99.2  &99.8  &\multicolumn{1}{l|}{99.5}  &91.2  &97.2   &\multicolumn{1}{l|}{92.1}  &99.6  &99.7  &\bf 99.7  \\ 
\MEAAD (Voting) &91.7   &91.7  &\multicolumn{1}{l|}{91.0}  &98.6  &98.6  &\multicolumn{1}{l|}{98.6}  &88.7  &88.7   &\multicolumn{1}{l|}{88.1}  &96.8  &96.8  &96.7  \\ 
\multicolumn{1}{l}{\MEAAD (Detector)}                    & \multicolumn{1}{l}{\bf 98.5} & \multicolumn{1}{l}{\bf 99.8} & \multicolumn{1}{l|}{\bf 98.6} & \multicolumn{1}{l}{\bf 99.6}    & \multicolumn{1}{l}{\bf 100}    & \multicolumn{1}{l|}{\bf 99.6}    & \bf 95.3 &\bf 99.2   &\multicolumn{1}{l|}{\bf 95.5}   &\bf 99.7   &\bf 99.7    &\bf 99.7         \\ 
\bottomrule
\end{tabular}
\end{threeparttable}
\label{table:sota}
\vspace{-0.4cm}
\end{table*}

\section{Experiments}
 
\subsection{Implementation details}
\noindent {\bf Datasets.} We validate the adversarial attack detection performance of \MEAAD on both Market1501 \cite{zheng2015market} and DukeMTMC-ReID \cite{ristani2016dukemtmc} datasets. Market1501 is captured by six cameras.
The training dataset contains 12,936 cropped images of 751 identities, while the testing set contains 19,732 cropped images of 750 identities.
DukeMTMC-ReID dataset is captured by eight cameras.
There are 16,522 bounding boxes of 702 identities for training and another 702 identities of 17,661 images for testing. We follow the standard training and testing splits for these two datasets in our experiments.

\noindent {\bf Attack implementations.}
We evaluate our defense strategy against two state-of-the-art adversarial attack approaches (Deep Mis-Ranking~\cite{wang2020deepmisrank} and advPattern~\cite{wang2019advpattern}) that are specifically designed against ReID systems, four attacks (FGSM \cite{goodfellow2015fgsm}, CW \cite{papernot2016cw}, Deepfool \cite{moosavi2016deepfool} and PGD \cite{madry2017pgd}) that are designed for general DNNs, and two adaptive attacks (adaptive CW and multi-model targeted attack) against \MEAAD.
The two attacks specific to ReID are described:

$\bullet$ {\it {Deep Mis-Ranking}} 
\cite{wang2020deepmisrank} is a digital attack that perturbs the ranking of the ReID system's outputs by proposing a learning-to-misrank formulation.

$\bullet$ {\it {advPattern}} \cite{wang2019advpattern} is a physical-world attack against ReID systems which adds printable adversarial patterns on clothes. 
Note that to do evaluation on a large scale, we do not print the generated patterns and add them physically. Instead, we add the patterns digitally onto the person images. This favors attackers since they can control how their physical perturbations are captured. 

\noindent {\bf Defense implementations.} 
Top-15 retrievals are used as the support set for each query.
To create an expert system with high heterogeneity, person ReID models with different network architectures are used during evaluation. Due to their superior performance on the Market1501 dataset, PCB \cite{sun2018pcb}, AlignedReID (AR) \cite{zhang2017alignedreid}, HACNN \cite{li2018hacnn}, LSRO \cite{zheng2017lsro} and Mudeep (MD)\cite{qian2017mudeep} 
are the five candidates to serve as expert models, and similarly, AlignedReID \cite{zhang2017alignedreid}, LSRO \cite{zheng2017lsro} HHL \cite{zhong2018hhl}, CamStyle (CS) \cite{zhong2018camerastyle} and SPGAN \cite{deng2018spgan} are the five candidates to serve as expert models for evaluation on the DukeMTMC-ReID dataset. 
For all the eight models, we use the author-released models with trained parameters. The ReID performance of the methods on both datasets is presented in the Supplementary Material.
The detector in \MEAAD is an MLP classifier with 2 hidden layers that contain 512 and 256 nodes respectively. ReLU function is used as the activation function. 
In addition to collecting benign context features for the detector training, we collect adversarial context features by perturbing the query samples in the training set with Deep Mis-Ranking~\cite{wang2020deepmisrank}, advPattern~\cite{wang2019advpattern} and other attacking methods. Note that there is no query/gallery separation in the training set; we randomly choose one person image as the query sample and use all the others as the gallery samples.
SGD optimizer with momentum 0.9 is used for training. The learning rate is 1e-4. The detector training is finished after 5,000 iterations and batch size is set to 1,024. Our experiments are conducted on a NVIDIA GTX 2080TI GPU using Pytorch.

\noindent {\bf Evaluation metric.} To tell if a query image input into the ReID system is perturbed, we first get the retrieval results from the chosen expert models and extract the context feature; the context feature is then input into the detector to be classified to attack or benign. Therefore, one metric used to evaluate the detection performance is the classification accuracy or called detection accuracy (\textit{Acc}). We keep the number of benign and perturbed samples balanced in our testing set. In addition to using probability threshold 0.5 to decide perturbed or not, we can flexibly adjust the threshold and get the Receiver Operating Characteristic (ROC) curve, for which, we report area under the ROC curve, i.e., \textit{ROC-AUC}, as another detection performance metric. Similarly, we also use \textit{F1 score} which is the harmonic mean of the dection precision and recall as one metric. 

\subsection{Attack detection performance}
In this section, we evaluate the proposed adversarial detection method against both Deep Mis-Ranking attack and advPattern attack on the Market1501 and DukeMTMC-ReID datasets. 
Three state-of-the-art detection methods
are extended to deal with ReID systems; they are used as the baseline methods which are described below. More details can be found in the Supplementary Material. 

 $\bullet$ \textit{Local Intrinsic Dimensionality (LID)} \cite{ma2018characterizing} characterizes the intrinsic dimensionality of adversarial regions which is a property of datasets \cite{amsaleg2015estimating}. Adversarial perturbation affects the LID characteristics of adversarial regions, and thus they are used to detect adversarial examples.

 $\bullet$ {\it Deep $k$-Nearest Neighbors (D$k$NN)} \cite{papernot2018deep} 
 combines the $k$-NN algorithm with feature representations of samples: an input is compared to its neighbors in the metric space. Labels of these neighbors afford confidence estimates for inputs outside the model’s training manifold, e.g. adversarial examples, which are used to detect adversarial attacks.
 
$\bullet$ {\it Spatial Rich Model (SRM)} \cite{fridrich2012rich,liu2019detection} detects adversarial examples from steganalysis point of view and proposes enhanced steganalysis features which are sensitive to small perturbations. Therefore, it can be used to distinguish the perturbed samples from the benign ones.

Moreover, instead of extracting the complete context feature and training a data-driven detector, we simply use the number of common support samples across all expert models as the feature and threshold over it to decide attack or benign. This is used as the forth baseline (\MEAAD(Voting)).

For the evaluation on Market1501, AlignedReID is chosen as the attack target model which is known to the attacker, and AlignedReID, LSRO, PCB and HACNN are used as the experts. 
For the evaluation on DukeMTMC-ReID, LSRO is the attack target model, and LSRO, SPGAN, AlignedReID and HHL are selected as the experts.

The detection performance is shown in Tab.~\ref{table:sota}.
We observe that advPattern attack is easier to be detected compared to Deep Mis-Ranking attack; the baseline methods and ours all have an F1 score over 96.7\% but ours (\MEAAD(Detector)) performs  better consistently. To detect the Deep Mis-Ranking attack, our method clearly outperforms the baseline methods on both datasets. For example, the F1 score on the Market1501 dataset of the D$k$NN method is 86.9\%; that of the LID method is 91.1\%; that of the SRM method is 94.1\%; that of \MEAAD(Voting) is 91.0\% with $threshold=5$; and \MEAAD(Detector) achieves 98.6\%, which is 4.5\% better than the best baseline. 

\begin{table}
\centering
\caption{Adversarial attack detection with different number of expert models on the Market1501 dataset. * indicates the attack target model known to the attackers.
}
\small
\setlength{\tabcolsep}{7pt}
\begin{threeparttable}
\begin{tabular}{llll}
\toprule
Expert models & Acc & AUC & F1 \\
\hline
AR* &95.2 &99.1 &95.5 \\
AR*+PCB &97.8 &99.7 &97.9 \\
AR*+PCB+LSRO &98.4 &99.8 &98.4\\
AR*+PCB+LSRO+HACNN     &98.5 &99.8 &98.6 \\

\bottomrule
\end{tabular}

\end{threeparttable}
\label{table:experts_NUM}
\end{table}

\begin{table}
\centering
\caption{Adversarial attack detection with/without using the attack target model as one of the expert models on Market1501. 
}
\small
\setlength{\tabcolsep}{12pt}
\begin{threeparttable}
\begin{tabular}{llll}
\toprule
Expert models & Acc & AUC & F1 \\
\hline
AR* &95.2 &99.1 &95.5 \\
AR*+PCB+LSRO+HACNN &\textbf{98.5} &\textbf{99.8} &\textbf{98.6}\\
PCB &88.2 &95.1 &88.7 \\
PCB+LSRO &93.7 &98.5 &93.9 \\
PCB+LSRO+HACNN &94.2 &98.5 & 94.2\\
\bottomrule
\end{tabular}
\end{threeparttable}
\label{table:experts_TARGET}
\vspace{-0.4cm}
\end{table}

\subsection{Ablation study}
In this section, we do ablation studies to understand  a) how the number of expert models affects the detection performance; b) whether the detection performance is sensitive to the different choices of expert models; c) how the size of the support set affects the detection performance; d) the importance of the three types of relations (query-support relation, support-support relation and cross-expert relation) in attack detection. Deep Mis-Ranking is used to attack AlignedReID model for the evaluation on Market1501.

\noindent \textbf{Number of expert models.}
In this section, we study whether more expert models improve the detection performance of \MEAAD on the Market1501 dataset. As shown in Tab.~\ref{table:experts_NUM}, with more expert models, the detection performance is better.
We suppose this is because more expert models bring more context information and thus the extracted context features are more discriminative between benign and perturbed samples. 
Note that when the number of expert models is one, there is no cross-expert affinity feature, only query-support affinity feature and support-support affinity feature are used, in which case, however, we still get very good performance: F1 score on Market1501 is 95.5\%.
Combining four expert models (AlignedReID, LSRO, PCB and HACNN), we achieve the best detection performance: 98.5\% detection accuracy on the Market1501 dataset.
The results on the DukeMTMC-ReID dataset can be found in the Supplementary Material.

\noindent\textbf{Choices of the expert models.}
In this section, we explore the detection performance of \MEAAD with different expert model choices.  Two choice strategies are compared, that is, including the attack target model as one of the expert models, and not using the attack target model as one of the expert models. 
The results are shown in Tab.~\ref{table:experts_TARGET}.
We observe that on the Market1501 dataset, the F1 score of using the attack target model (AR model) as the only expert model is 95.5\%, which is higher than 94.2\% when using other three expert models (PCB+LSRO+HACNN). This indicates that it is beneficial to include the attack target model as one of the expert models. 
The same conclusion can be drawn on the DukeMTMC-ReID dataset and the details can be found in the Supplementary Material.
A potential reason is that the attack is tuned to the target model and this creates a larger variance with the other experts in the retrievals. 

\noindent \textbf{Size of the support set.}
Note that all the previous evaluations are done with the size of support set equal to 15, basically top-15 retrievals are used to extract the context. We explore how the size of the support set affects the attack detection performance. Specially, we evaluate the attack detection performance when $K=1,5,10,15,20,30$ as shown in Tab.~\ref{table:topk}.
Note that $K=1$ means there is no support-support affinity feature, and only query-support affinity feature and cross-expert affinity feature are used. We observe that in general using a larger support set gives better attack detection rate, when $K=15$, we achieve 98.5\% detection accuracy - 6.2\% improvement comparing to the result of $K=1$. It can be seen that thereafter ($K>15$), with the increase of the support samples, the performance is almost stable. 

\begin{table}[]
\centering
\caption{Adversarial attack detection with different sizes of the support set on the Market1501 dataset.
}
\small
\setlength{\tabcolsep}{20pt}
\begin{threeparttable}
\begin{tabular}{llll}
\toprule
Top-$K$ & Acc & AUC & F1 \\
\hline
$K=1$ &92.3 & 99.2 &92.9 \\
$K=5$ &94.4 &99.7 &94.7\\
$K=10$ &97.5 &99.8 &97.6\\
$K=15$ &98.5 &99.8 &98.6 \\
$K=20$ & 98.5 &99.8 &98.5 \\
$K=30$ & 98.5 &99.8 &98.6 \\
\bottomrule
\end{tabular}
\end{threeparttable}
\label{table:topk}
\end{table}

\begin{table}
\centering
\caption{Ablation test: adversarial attack detection with different context features on the Market1501 dataset.
}
\small
\setlength{\tabcolsep}{10pt}
\begin{threeparttable}
\begin{tabular}{llllll}
\toprule
$A_{c-e}$ &$A_{q-s}$ &$A_{s-s}$ & Acc & AUC & F1 \\
\hline
\checkmark & &  &94.9 &99.4 &95.1\\ \hline
&\checkmark & &93.6 &99.6 &94.0\\ \hline
& &\checkmark &90.9 &97.8 &91.6\\ \hline
\checkmark &\checkmark & &97.2 &99.5 &97.1\\ \hline
\checkmark & & \checkmark&95.6 &99.6 &95.7\\ \hline
&\checkmark &\checkmark &97.0 &99.5 &97.0\\ \hline
\checkmark & \checkmark & \checkmark&\textbf{98.5} &\textbf{99.8} &\textbf{98.6}\\
\bottomrule
\end{tabular}
\end{threeparttable}
\label{table:ablation}
\vspace{-0.4cm}
\end{table}

\begin{table*}[]
\centering
\caption{Adversarial attack detection against unseen new attacks on the Market1501 dataset.
}
\small
\begin{tabular}{llllllllllllll}
\toprule
\multirow{3}{*}{Settings} & \multirow{3}{*}{Attacks} & \multicolumn{12}{c}{Testing}                                                                                  \\ \cline{3-14} 
                          &            & \multicolumn{3}{c}{CW} & \multicolumn{3}{c}{Deepfool} & \multicolumn{3}{c}{FGSM} & \multicolumn{3}{c}{PGD} \\ \cline{3-14} 
                          &            & Acc   & AUC    & F1    & Acc     & AUC      & F1      & Acc    & AUC    & F1     & Acc   & AUC    & F1     \\ \cline{1-14} 
\multirow{4}{*}{Training} & CW         &\cellcolor[HTML]{EFEFEF}96.1 &\cellcolor[HTML]{EFEFEF}98.7 &\cellcolor[HTML]{EFEFEF}96.2  & 96.1     & 98.6     & 96.2    & 96.2    & 98.6   & 96.3   & 96.9   & 99.0   & 97.0   \\ 
                          & Deepfool   & 95.8   & 98.6   & 95.9  &\cellcolor[HTML]{EFEFEF}96.0    &\cellcolor[HTML]{EFEFEF}98.6     &\cellcolor[HTML]{EFEFEF}96.1    & 95.8    & 98.6   & 95.9   & 96.6   & 98.9   & 96.7   \\ 
                          & FGSM       & 96.1   & 98.7   & 96.2  & 96.0     & 98.7     & 96.1    &\cellcolor[HTML]{EFEFEF}96.1    &\cellcolor[HTML]{EFEFEF}98.8   &\cellcolor[HTML]{EFEFEF}96.2   & 96.8   & 99.0   & 96.9   \\ 
                          & PGD        & 94.0   & 98.4   & 93.8  & 93.7     & 98.3     & 93.6    & 94.1    & 98.4   & 93.9   &\cellcolor[HTML]{EFEFEF}97.4   &\cellcolor[HTML]{EFEFEF}98.8   &\cellcolor[HTML]{EFEFEF}97.4   \\ \bottomrule
\end{tabular}
\label{table:attacking}
\vspace{-0.4cm}
\end{table*}

\noindent \textbf{Importance of different relations.}
The context feature used in \MEAAD is composed  of three parts, that is, query-support affinity feature  ($A_{q-s}$), support-support affinity feature  ($A_{s-s}$) and cross-expert affinity feature  ($A_{c-e}$).
To quantify the contribution of each relation, we conduct an ablation study on the Market1501 dataset. As shown in Tab.~\ref{table:ablation}, when using only one relation for detecting attacks, we have already achieved over 90.5\% detection accuracy in all cases, especially 94.9\% for cross-expert affinity. 
All the three features are complementary, that is, combining two of them improves the detection performance consistently, such as the detection accuracy is increased from 90.9\% (support-support affinity) to 97.2\% (cross-expert affinity and query-support affinity). When using all of them, our method achieves the best attack detection performance 98.5\% for F1 score, 99.8\% AUC score and 98.6\% detection accuracy.

\subsection{Defense against new attack methods} 
\label{other attacking}
Recall that we need both the benign and perturbed samples to train our detector in \MEAAD. Therefore, the detector is able to detect the attacks which have appeared in its training set. However, as new attack methods are proposed, it is not feasible to exhaustively cover all the attack methods in the training set. In this section, we evaluate how our defense method transfers to new unknown attacks. 
We extend four state-of-the-art adversarial attack methods against general DNNs to attack ReID systems; basically, we regard the last layer of the ReID model as the identity prediction layer, and on top of that the adversarial query examples can be generated for training and testing. The four attack methods: FGSM \cite{goodfellow2015fgsm}, CW \cite{papernot2016cw}, Deepfool \cite{moosavi2016deepfool} and PGD \cite{madry2017pgd} are implemented with Torchattacks \cite{kim2020torchattacks}.

We generate the training set with one attack method, and test the trained detector on all the attack methods. The experiment is done on the Market1501 dataset.
AlignedReID is used as the attack target model. AlignedReID, LSRO and PCB are used as the three expert models. The results are shown in  Tab.~\ref{table:attacking}.  When the detector is tested on the same attack method as that used in its training set, the detection performance          is very good, for example, F1 score equals to 96.2\% when detecting CW attack. When the detector is asked to detect unseen attacks, the performance remains or drops just by a little bit, for example, F1 score drops from 96.2\% to 95.9\% when detecting unknown Deepfool attack compared to known CW attack. 
This indicates that although different attack methods generate the perturbations in different ways, the perturbations tend to always cause messed up retrieval results, and thus our defense strategy of detecting context inconsistency transfers well across different attack methods and is effective to unknown new attacks.

\subsection{Adaptive Attacks against \MEAAD}
\label{adaptive attack}
To further evaluate \MEAAD's robustness towards adaptive attacks, we extend an existing adaptive attack method, i.e., adaptive CW attack, and adopt a new adaptive attack method, i.e., multi-model targeted attack to evade \MEAAD. More details can be found in the Supplementary Material.

\noindent \textbf {Adaptive CW attack.} 
We extend the adaptive CW algorithm \cite{carlini2017adversarial} by
introducing a new loss item (associated with three kinds of context defined in \MEAAD) to the loss function and the new loss term is as below: 
\begin{equation}
    l_{*}(\MEAAD(x_{adv}))=-\sum(A_{qs}+A_{ss}+A_{ce})
    \label{eq:MEAAD_score}
\end{equation}
The new item is defined to reduce the retrieval inconsistency of the adversarial examples. The rationale for the minimization of the added term in Eq.~\ref{eq:MEAAD_score} is that adversarial examples have lower context affinity than benign examples.
Experiments show that such adaptive attack decreases \MEAAD's detection accuracy by 1.3\% when only the attack target ReID model is white-box to the attacker, and 3.5\% when all the ReID models are white-box to the attacker. In either way, the  ROC-AUC score of \MEAAD is still high, over 95\%. 

\noindent \textbf {Multi-model targeted attack.} As shown in Fig.~\ref{fig:fig1}, the retrieval results of non-targeted attack are messy and not consistent across different models, and thus such attacks are detected by \MEAAD. If we assume all expert ReID models are white-box to the attacker, then the attacker could do targeted attack against all the models simultaneously. In other words, this adaptive attack generates adversarial examples that fool all the ReID models used in \MEAAD (both the target model and the expert models) to retrieve the same wrong identity and thus context would be more consistent.
We extend the adversarial metric attack in \cite{bai2020adversarial} to a multi-model targeted attack for attacking \MEAAD.  
However, aligned with previous works~\cite{zhu2020you}, we find that targeted attack against multiple ReID models is hard and only 211 (6.2\%) such adversarial examples from all the 3,368 testing samples. \MEAAD's detection accuracy on the 211 adversarial examples is 88.6\%.



\section{Conclusions}

In this paper, we propose a Multi-Expert Adversarial Attack Detection framework that detects adversarial attacks against ReID systems by checking context inconsistency, a side effect of the adversarial attacks.
Empirical studies show that query-support affinity, support-support affinity and cross-expert affinity 
are able to distinguish the perturbed ones. Therefore, we propose to leverage the three relations to form the context feature for each query sample. A detector is then trained on the context features of both benign and perturbed samples and are then used to detect adversarial attacks.
Experiments on the Market1501 and DukeMTMC-ReID datasets show that \MEAAD effectively detects various adversarial attacks, that is, Deep Mis-Ranking, advPattern, Deepfool, CW, FGSM and PGD, and \MEAAD can effectively detect unknown new attacks. 

\textbf{Acknowledgments.}
This work was supported in part by the National Natural Science Foundation of China under Grant 62073126 and Grant 61771189, in part by the Hunan Provincial Natural Science Foundation of China under Grant 2020JJ2008, and in part by the Office of Naval Research (ONR) at UCR under Grant N00014-19-1-2264.




\small
\bibliographystyle{ieee_fullname}
\bibliography{egbib}
\end{document}